\documentclass{article} 
\usepackage{iclr2025_conference,times}

\usepackage{amsmath,amsfonts,bm}

\def\eqref#1{equation~\ref{#1}}

\def\1{\bm{1}}

\DeclareMathAlphabet{\mathsfit}{\encodingdefault}{\sfdefault}{m}{sl}
\SetMathAlphabet{\mathsfit}{bold}{\encodingdefault}{\sfdefault}{bx}{n}

\usepackage{hyperref}
\usepackage{url}
\usepackage{graphicx}
\usepackage{soul} 
\usepackage{booktabs} 

\title{Decomposing Irregular Time Series into Interpretable Temporal Patterns with Mixture-of-Experts}

\author{Antiquus S.~Hippocampus, Natalia Cerebro \& Amelie P. Amygdale \thanks{ Use footnote for providing further information
about author (webpage, alternative address)---\emph{not} for acknowledging
funding agencies.  Funding acknowledgements go at the end of the paper.} \\
Department of Computer Science\\
Cranberry-Lemon University\\
Pittsburgh, PA 15213, USA \\
\texttt{\{hippo,brain,jen\}@cs.cranberry-lemon.edu} \\
\And
Ji Q. Ren \& Yevgeny LeNet \\
Department of Computational Neuroscience \\
University of the Witwatersrand \\
Joburg, South Africa \\
\texttt{\{robot,net\}@wits.ac.za} \\
\AND
Coauthor \\
Affiliation \\
Address \\
\texttt{email}
}

\begin{document}

\maketitle

\begin{abstract}
Irregularly-spaced sequential data, such as transaction logs, encode multi-scale temporal structure that is critical for prediction but difficult to recover explicitly with standard sequence models, which often trade interpretability for performance. We introduce MoTIF, an interpretable framework using mixture-of-experts (MoE) attention to decompose irregular time series into temporal patterns spanning minute-to-year scales. Each expert is trained via temporal alignment that induces attention specialization to a designated cyclical or within-range pattern, with label-free routing supervision that learns to activate the correct expert from patterns present in the data. On synthetic data with known temporal structure, MoTIF recovers correct routing well above chance (0.389/0.829 for range/cyclical patterns vs. 0.091 random) and outperforms all baselines, including a standard transformer and an unspecialized MoE, on next-value prediction. On a public credit card transaction dataset, MoTIF achieves higher macro-F1 than all baselines for next-amount prediction. Its learned routing reveals structure a black-box model cannot: transactions are governed primarily by intraday and short-interval rhythms, while fraudulent transactions show bursty, calendar-agnostic timing versus the steadier periodicity of legitimate ones. On downstream fraud detection, MoTIF matches standard transformer and MoE baselines and clearly outperforms recent sequential recommendation models used as embedding baselines. These results show that decomposing temporal structure into interpretable, expert-specific patterns need not sacrifice predictive accuracy, offering a practical route to sequence models that are both accurate and auditable in high-stakes settings. Code is available at \url{https://anonymous.4open.science/r/MoTIF-82E4}.
\end{abstract}

\section{Introduction}
Forecasting and inference over temporal data underpins critical decision-making across domains such as healthcare, finance, and climate science \citep{emmert2026time,cao2026automatic}. Real-world observations, however, are frequently irregularly sampled, with non-uniform intervals and sparse measurements. While deep forecasting models such as transformers and RNNs excel under regular-grid assumptions, they struggle inherently with irregularity. Recent advances have substantially improved irregular time series modeling \citep{chen2023contiformer,zhang2023warpformer,park2024efficient,zhong2025mtm,benassi2025forecasting,lee2026adaptive}, yet these methods offer little insight into why a prediction was made, relying on opaque latent representations. In high-stakes settings, this opacity is a critical bottleneck: end-users cannot verify whether a forecast reflects stable trends, periodic shifts, or transient irregularities. Addressing it is paramount, as interpretability builds data-backed user trust \citep{yan2023self}, satisfies regulatory compliance standards \citep{arrieta2020explainable}, and helps domain experts extract actionable knowledge from complex temporal patterns \citep{bai2018interpretable}.

Human behaviors, including spending, media use, and health rituals, are largely driven by biological and social rhythms following daily, weekly, and seasonal patterns \citep{kreitzman2011rhythms,roenneberg2003life,southerton2020socio,hohm2024homo}, which encode underlying behavioral characteristics indicative of future actions \citep{beck2022personalized}. Such patterns can reveal anomalous behaviors like fraud while providing transparent, verifiable predictions \citep{darwish2025intelligent,hashemi2025time}.

Recent architectures modeling complex temporal patterns still fall short of granular interpretability and typically struggle with irregular data. Time-MoE \citep{shi2025timemoe} and Seg-MoE \citep{ortigossa2026seg} use mixture-of-experts routing over horizons or segments to capture multi-scale behavior, but do not isolate the specific temporal patterns detected. MoHETS \citep{ortigossa2026mohets} uses Fourier experts to separate noise from trend but lacks structural interpretability, compressing long-term patterns into a general expert. Data-driven routing approaches like RoMA and STAR \citep{li2026routing,park2026star} align routing with sample similarity but do not reveal the temporal patterns underlying that alignment. AME-TS \citep{wang2026amets} incorporates seasonality and trend descriptors into a soft routing prior, but its supervision is sequence-level only, and evaluation focuses on forecasting error without auditing expert semantics. Critically, none of these architectures is designed for irregularly sampled data, as their reliance on regular grids or spectral transforms limits direct application to irregular gaps.

To address these limitations, we introduce MoTIF (Mixture of Temporal Interpretable Function-experts), an interpretable prediction framework for irregular time series that uses mixture-of-experts attention to explicitly decompose predictive contributions by temporal pattern. Our contributions are:
(1) Specialized experts explicitly capture temporal patterns across minute-to-year scales by aligning attention with cyclical and range-based autoregressive trends.
(2) A label-free routing-supervision scheme recovers expected routing on synthetic data with known pattern labels; on real transaction data, top-routed experts align with detected temporal structure, letting routing weights serve as a pattern signature without training labels.
(3) The method improves next-value prediction on synthetic data and maintains multi-task next-token performance and downstream fraud-prediction generalizability on real data, adding interpretability without sacrificing accuracy.
(4) Analysis of the temporal patterns driving next-token prediction reveals transaction-, sequence-, and dataset-level insights, including systematic fraud/non-fraud differences consistent with fraud's burstier, calendar-agnostic timing versus legitimate transactions' steadier periodicity.

\section{Related Work}

\textbf{Representation Learning for Sequential Data.}
Transaction logs are naturally represented as \emph{marked event sequences}, $(t_i,\mathbf{x}_i)_{i=1}^{N}$, with informative timestamps and attributes. Sequential recommenders are closely related: SASRec applies causal self-attention to interaction histories \citep{kang2018sasrec}; TiSASRec adds positional and pairwise time-interval encodings \citep{li2020tisasrec}; D{\'e}j{\`a} vu modulates attention via context-selected temporal kernels \citep{wu2020dejavu}; and HSTU/PANTHER scale generative pretraining to heterogeneous action sequences, with PANTHER also modeling periodic motifs \citep{zhai2024hstu,li2026panther}.

Closer to our setting, TabFormer pretrains on multivariate tabular sequences using the same simulated credit-card corpus used here \citep{padhi2021tabformer}, and CoLES learns contrastive embeddings from financial event sequences \citep{babaev2022coles}. Both are strong generic encoders but represent temporal effects only via features, interval embeddings, or unconstrained kernels. MoTIF instead assigns fixed, interpretable meanings to its experts, exposing at each event the relative support for a predefined calendar-cycle or elapsed-range pattern.

\textbf{Interpretable Mixture-of-Experts.}
Mixture-of-experts models add capacity via routing \citep{shazeer2017outrageously}, but specialization alone does not yield stable expert semantics. In sequential recommendation, FAME uses expertized attention for preference facets \citep{liu2025fame}, HM4SR routes temporal experts via timestamp embeddings \citep{zhang2025hm4sr}, and FamouSRec routes users by behavioral frequency \citep{zhang2025famousrec}. Among time-series foundation models, Time-MoE uses sparse experts mainly for scalability \citep{shi2025timemoe}, while Moirai-MoE learns automatic token-level specialization \citep{liu2025moiraimoe}; other architectures couple MoE routing with anomaly detection \citep{zhang2026act} or route attention/feed-forward paths independently \citep{yang2026experttransformer}. These methods yield useful heterogeneity, but expert meaning is typically discovered post hoc or tied to architecture rather than to a fixed, human-readable temporal concept.

Interpretable MoE work makes expert roles explicit: Interpretable Mixture of Experts pairs hard assignment with simple predictors for exact local explanations \citep{ismail2023interpretablemoe}, and I$^2$MoE weakly supervises experts toward predefined multimodal interaction types \citep{xin2025i2moe}. MoTIF applies an analogous principle to irregular event sequences: each expert is aligned with a named calendar-cycle or elapsed-range kernel, supervised by a detached, label-free teacher via predictive and autocorrelation evidence rather than ground-truth annotations. Since attention and gating weights are not inherently faithful explanations \citep{jain2019attention,wiegreffe2019attention}, we treat routing as a structural diagnostic rather than a causal one, testing whether experts match prescribed kernels, whether synthetic patterns recover the corresponding experts, and whether held-out routing agrees with directly computed temporal statistics.

\section{Method}
Prior work on irregular time series modeling often lacks interpretability. We propose explicitly decomposing irregular sequences into cyclical and elapsed-range temporal patterns at human-relevant scales, from minute to year. The preliminary background on
temporal pattern decomposition is provided in Appendix Section \ref{appendix:preliminaries}.

\begin{figure}[t]
\begin{center}
\includegraphics[width=\textwidth]{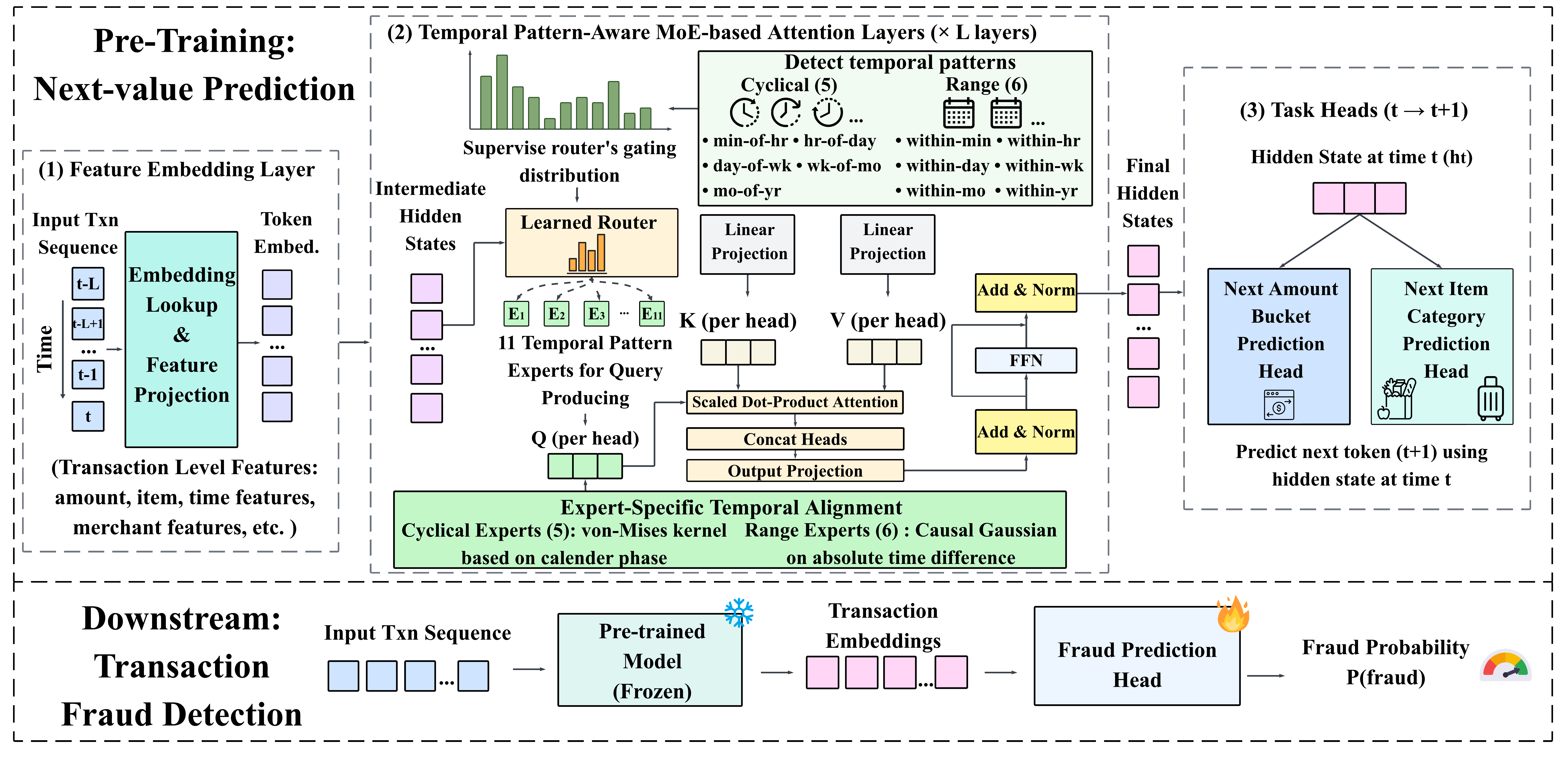}
\end{center}
\caption{\textbf{Method Schematic.} MoTIF consists of (1) a feature embedding layer, (2) a transformer backbone with temporal pattern-aware MoE-based attention query and label-free routing supervision, and (3) task heads. Continuous and categorical features are embedded via an MLP and learned lookup tables, respectively, and concatenated into a token embedding. These embeddings pass through a transformer whose self-attention query is computed by an MoE layer, where specialized experts induce attention patterns aligned with cyclical or elapsed-range temporal structure from minute to year scales. Label-free routing supervision trains the model to route sequences to experts matching their constituent patterns, based on next-value prediction performance and detected autocorrelations. Final representations feed into task heads for next-value prediction, and their generalizability is evaluated via DCN and GBT probes for downstream fraud detection.}\label{method_schematic} 
\end{figure}

We present MoTIF (Mixture of Temporal Interpretable Function-experts), comprising: (1) a feature embedding layer, (2) a transformer backbone with temporal-pattern-aware MoE attention, and (3) task heads for next-value prediction (Fig. \ref{method_schematic}).

The embedding layer is an MLP that generates initial feature representations at each time point; categorical variables use a lookup table of learned embeddings, concatenated with continuous-feature representations. These embeddings feed into a causally masked transformer for autoregressive next-token prediction. Within this transformer, we introduce a multi-head attention mechanism that uses MoE to decompose each sequence into canonical temporal patterns, replacing the standard query matrix with an MoE layer: each ``attention expert" produces one attention output per head, mixed by a router. Experts specialize in cyclical and range-based patterns reflective of human behaviors \citep{barabasi2005origin,roenneberg2003life} across minute, hour, day, week, month, and year scales \citep{monsa2020processing,schneider2013unravelling,sparks2022shifting,xu2020finding}, with the router upweighting whichever experts' patterns are most relevant to each sequence.

Attention specialization is learned via a temporal alignment loss that minimizes the cross-entropy between an expert's attention and a target kernel modeling its assigned pattern. Since true pattern labels are rarely available in real sequential data, we introduce label-free routing supervision that upweights the appropriate experts per sequence based on each target kernel's autoregressive prediction signal and detected autocorrelations in the data. Together, alignment makes each expert an attention specialist for a given pattern type and scale, while routing supervision ensures the MoE selects the correct specialist per token.

\subsection{Temporal Pattern-Based Attention Expert Specialization}\label{sec:temporal_alignment}
Training each expert's attention matrix to attend to specific parts of the sequence teaches that expert to specialize in a known temporal pattern. We specialize experts toward two pattern types: cyclical and range-based. Each expert's attention distribution is aligned to an analytic target kernel encoding its assigned pattern via a temporal alignment loss. We construct these expert-specific target attention matrices per-sequence as follows:

(1) \textit{Cyclical experts:} Cyclical experts encourage the model to attend to time points that occur at the same recurring position in time, i.e. minute-of-hour, hour-of-day, day-of-week, week-of-month, month-of-year. We specify cyclical attention patterns using a von-Mises kernel $K^{\text{cyc}} \in \mathbb{R}^{T \times T}$ (\cite{taylor2008automatic}) based on the calendar phase. At each pair of time indices $i,j \in \{1,...,T\}$, for expert $e$, phase group $g$ (e.g. hour of day from 0-23), period $P$, and concentration $\kappa_e$:
\begin{equation}
    K^{\text{cyc}}_{e,{ij}} = 
  \exp\big(\kappa_e(\cos\theta_{ij} - 1)\big), \qquad \theta_{ij} = \frac{2\pi}{P}(g_i - g_j)
\end{equation}
Note that when $i,j$ share the same phase, the kernel peaks at 1, and it decays as the phase separation grows. $\kappa_e$ is the bandwidth and is set to 4 in our experiments.

(2) \textit{Range-based experts.} Range-based experts detect dependencies within a specified time range (minute, hour, day, week, month, year) by attending to nearby time points at that scale. We define the target attention matrix $K^{\text{rng}} \in \mathbb{R}^{T\times T}$ as a causal Gaussian over absolute time difference: for expert $e$ with characteristic scale $s_e$, and timestamps $i,j$ represented in cumulative hours $h_i, h_j$ since the first transaction,
\begin{equation}
K^{\text{rng}}_{e,ij} =
\exp\left(-\tfrac{1}{2}\Big(\frac{\Delta{ij}}{s_e w_e}\Big)^{2}\right), \qquad \Delta_{ij} = h_i - h_j
\end{equation}
where $s_e$ is the expert-specific range in hours (e.g., 24 for within-day, matching the units of $h$) and $w_e$ is the bandwidth multiplier (set to 1 in our experiments). The product $s_e w_e$ controls how quickly the influence of time $j$ on time $i$ decays as their lag grows.

For all temporal patterns, the expert-specific target kernels are causally masked and normalized to have unit row sums, reflecting proper distributions over the time sequence at each query.


\textbf{Cross-entropy temporal alignment loss.} We minimize cross-entropy loss between the target and actual attention matrices to encourage the expert-specific attention to specialize for the designated temporal patterns:
\begin{equation}
    \mathcal{L}_{\text{align}} = \frac{1}{|E|}\sum_{e \in E} \mathcal{L}_{\text{align},e} \quad\quad\quad \mathcal{L}_{\text{align},e} = -\frac{\sum_{i,j} t_{e,ij}\log\big(a_{e,ij}\big)\cdot 
  \text{M}_{ij}}{\sum_{i,j}\text{M}_{ij}}
\end{equation}
where $M_{ij} \in \{0,1\}$ is a mask matrix with $M_{ij} = 1$ for $j \leq i$ and $j$ not padding, and $t_{e,ij}$ and $a_{e,ij}$ are the target and actual computed attention matrix values for expert $e$ between time points $i$ and $j$, respectively. Note that $a_{e,ij}$ is computed from expert $e$'s standalone query, $q_{e,i}$, for this loss to induce expert-specific attention patterns. The final alignment loss $\mathcal{L}_{\text{align}}$ is averaged over the experts.

\textbf{Multi-head MoE-based query attention.} This is implemented within a \textit{multi-head} framework. For each head, the MoE component produces a query vector at each time point as a router-gated mixture of expert-specific queries. Keys and values, attention within each head, and the final multi-head projection all follow the standard formulation (scaled dot-product attention with softmax, followed by concatenation across heads and a learned output projection), producing position-wise representations. These are aggregated into a sequence-level embedding via standard attention pooling: a learnable vector scores each position, the scores are softmax-normalized into weights, and the final embedding is their weighted sum over positions. More explicit formulation details can be found in the Appendix (Section \ref{appendix:multihead_query_attn}).


\subsection{Label-free Routing Supervision}\label{sec:routing_supervision}
Temporally aligning the attention matrices induces expert specialization, but does not guarantee that sequences are routed to the experts matching their predominant patterns, especially since the true temporal patterns present in each sequence are typically unknown and cannot directly supervise the router. We therefore propose a \textit{label-free routing supervision} mechanism that uses a predictive-responsibility Kullback-Leibler (KL) divergence loss to push routing toward temporal pattern-aware experts, based on predictive performance and autocorrelation detected in the data. This mechanism has two steps: (1) build a detached teacher, and (2) score the router against it with a decoupled tri-term loss.

\textbf{Detached teacher.} The detached teacher scores experts hierarchically using: (a) predictive error, (b) pattern family relevance (cyclical vs. range-based), and (c) within-family relevance (which specific scale is most relevant within the identified family).

\textit{Per-expert predictive error.} Each specialized expert $e$ forecasts the next value at position $i$ from strictly-past time points, weighted by the (rather than the currently-optimized attention weights, to reduce training noise) target attention kernel:
\begin{align}
    &\epsilon_{e,i} = (\hat{y}_{e,i} - x_i)^2, \qquad
    \hat{y}_{e,i} = \sum_j K_{\text{tgt-past},e,ij} \cdot x_j, \qquad K_{\text{tgt-past},e,ij} = \frac{K_{e,ij} \cdot \mathbf{1}_{j < i}}{\sum_{k} K_{e,ik} \cdot \mathbf{1}_{k < i}}
\end{align}
where $i,j$ are time indices, $K_{\text{tgt-past},e}$ is the causally-masked, re-normalized target kernel for expert $e$, $\mathbf{x} = [x_1,...,x_T]$ is the input sequence, and $\hat{y}_{e,i}$ is expert $e$'s prediction (the autoregressive label being $x_i$). Thus $\epsilon_{e,i}$ measures how well expert $e$'s pattern and scale predict each token; this is the raw signal all subsequent teacher stages transform. Because this assumes a linear relationship between relative time points, it recovers cyclical patterns fairly well but only partially discriminates range patterns and their scales, motivating the following stages.

\textit{Stage 1: Pattern family split via cyclical-first gate.} Since the error signal from above resolves cyclical experts well but is dominated by them, we add a cyclical-first gate to first isolate pattern family (cyclical vs. range) before resolving within-family signals:
\begin{align}
    &[P_{i,\text{range}}, P_{i,\text{cyclical}}] = \text{Softmax}\left(\frac{[\theta_{thr},r^2_{i,cyc}]}{\tau}\right), \qquad
    r^2_{i,cyc} = 1-\frac{\min_{e \in {E_{cyc}}}\epsilon_{e,i}}{V_i}
\end{align}
where $E_{cyc}$ is the set of all cyclical experts, $V_i$ is the variance of past values preceding time $i$ (clamped to $10^{-8}$ to avoid division by zero, and decoupling the family decision from absolute error scale), $\theta_{thr}$ is the error-based decision boundary ($\theta_{thr} = 0$), and $\tau$ is the temperature parameter ($\tau = 1$). Higher cyclical weights are assigned when the best cyclical expert explains the token much better than the mean.

\textit{Stage 2: Within-family resolution.} Within each family, the per-expert probability is:
\begin{equation}
    r = P(\text{family} | \text{cyc-first gate}) \times P(\text{expert} | \text{family})
\end{equation}
where $P(\text{family}|\cdot) \in \{P_{\text{range}}, P_{\text{cyclical}}\}$ from Stage 1, and $P(\text{expert}|\text{family})$ is given by the within-cyclical ($q_{cyc}$) or within-range ($q_{rng}$) distributions below.

\textbf{Within-cyclical resolution via error signal.} We re-normalize the per-expert predictive errors over the cyclical experts:
\begin{align}
    &{q_{cyc}} = \text{Norm}([\rho_{err,e} | e \in E_{cyc}]), \qquad
    \rho_{err} = \text{Softmax}\left(-\frac{[\epsilon_e | e \in E]}{\tau}\right)
\end{align}
where $E$ is the set of all experts, $\epsilon_e$ is the per-expert error, $\tau = 1$, and Norm($v$) normalizes $v$ to sum to 1.

\textbf{Within-range resolution via structure function.} Since range experts detect autocorrelation at scale $s_e$, we assume the correct scale lies within the autocorrelation decay length, resolved by fitting each expert's known-$s$ Ornstein-Uhlenbeck (OU) structure function \citep{uhlenbeck1930theory} $S_e$ to the empirical squared increments $u$:
\begin{align}
    S_{e,ij} = 2V\left(1-\exp\left(-\frac{|t_i - t_j|}{s_e}\right)\right), \qquad u_{ij} = (x_i - x_j)^2
\end{align}
where $V$ is the per-sequence empirical variance of the value signal (serving as amplitude) and $x_i$ is the value at time $i$. The fit error $m_{e,i}$ determines the within-range probability $q_{rng}$:
\begin{equation}
    m_{e,i} = \frac{1}{|T_{j<i}|}\sum_{t \in T_{j<i}} (u_{ij} - S_{e,ij})^2, \qquad {q_{rng}} = \text{Softmax}\left(\frac{Z([-m_e | e \in E_{rng}])}{\tau}\right)
\end{equation}
where $Z(\cdot)$ is z-score normalization, $E_{rng}$ is the set of range-based experts, and $\tau = 1$. The per-expert probabilities are then ${r_{cyc}} = P_{\text{cyclical}} \cdot {q_{cyc}}$ and ${r_{rng}} = P_{\text{range}}  \cdot {q_{rng}}$, yielding the detached teacher ${r} = [{r_{cyc}}, {r_{rng}}]$ toward which the router is trained.

\textbf{Router supervision loss: decoupled tri-term KL.} We construct the router supervision objective as three independently-weighted, per-family-normalized KL terms, decoupling family gating from within-family assignment to allow fine-grained control and avoid interference between the two. For each family $f \in \{cyc, rng\}$, we split the teacher into family mass $m_f$ and within-family target $q_f$:
\begin{align}
    &m_{f} = \sum_{e \in E_f} r_{f,e}, \qquad {q_{f}} = \frac{{r_{f}}}{m_{f}} 
\end{align}
and similarly split and renormalize the router's expert probabilities $p$:
\begin{align}
    &p_{f,fam} = \sum_{e\in E_f} p_{f,e} \quad\quad {p_{f,norm}} = \frac{{p_{f}}}{p_{f,fam}}
\end{align}
The family-level loss is the KL divergence between the teacher and router family-mass vectors $m_i=[m_{cyc,i},m_{rng,i}]$ and $p_{fam,i}=[p_{cyc,fam,i},p_{rng,fam,i}]$:
\begin{align}
    &\mathcal{L}_{fam} = \frac{1}{N}\sum_{i=1}^N \text{KL}(m_i || p_{fam,i})
\end{align}
and the within-family loss for $f \in \{cyc,rng\}$ is the mass-weighted KL divergence between within-family teacher and router distributions:
\begin{align}
    &\mathcal{L}_{f} = \frac{\sum_i m_{f,i}\cdot \text{KL}({q_{f,i}} || {p_{f,norm,i}})}{\sum_j m_{f,j}\cdot n_{heads}}
\end{align}
The total routing supervision objective combines all three terms:
\begin{equation}
    \mathcal{L}_{route} = w_{fam}\mathcal{L}_{fam} + w_{cyc}\mathcal{L}_{cyc} + w_{rng}\mathcal{L}_{rng}
\end{equation}

\subsection{Training Procedure} 
MoTIF is trained in two phases. In Phase 1, we pretrain the model on next-token prediction with standard autoregressive cross-entropy loss, jointly optimized with our temporal alignment loss and label-free routing supervision loss (Sections~\ref{sec:temporal_alignment} and~\ref{sec:routing_supervision}), plus a load-balancing auxiliary loss \citep{shazeer2017outrageously} to prevent expert collapse. This phase yields representations that are both predictive and temporally interpretable, as shown in Section~\ref{sec:results}.

In Phase 2, we fine-tune the pretrained backbone with a Deep \& Cross Network (DCN) head \citep{wang2017deep} for fraud prediction, trained with standard binary cross-entropy loss. To verify that the pretrained representations carry fraud-relevant signal independent of this task-specific head, we additionally probe them with an independent gradient-boosted tree classifier \citep{friedman2001greedy}.

Full loss formulations, weighting terms, and hyperparameters for both phases are provided in Appendix~\ref{appendix:training_details} and \ref{appendix:impl_details}.

\section{Experimental Results} \label{sec:results}
Our pipeline consists of an autoregressive next-token pretraining phase (Phase 1), which yields representations that predict future values while capturing underlying temporal patterns, and a downstream fine-tuning phase (Phase 2) that tests their generalizability; here, we evaluate on fraud prediction. Training details are in the Appendix (Section \ref{appendix:impl_details}). We first validate interpretability and performance on a synthetic dataset spanning diverse temporal patterns, then apply our approach to a public credit card transaction (CCT) dataset \citep{cct}, using MoTIF's explainability feature to analyze its temporal structure (dataset details in Sections \ref{appendix:synthetic} and \ref{appendix:cct}). We benchmark MoTIF against a plain transformer and an unspecialized attention-MoE baseline (FAME) \citep{vaswani2017attention,liu2025fame}, along with prior methods for irregular sequential prediction: PANTHER, HSTU, DuoRec, and SASRec \citep{li2026panther,zhai2024hstu,duorec,kang2018sasrec}. We assess both interpretability (attention specialization and routing) and predictive performance; evaluation metric details are in Section \ref{appendix:eval_metrics}.

\subsection{Interpretability and Predictive Gains on Synthetic Data}

We assess whether MoTIF learns temporal pattern-specialized attention and routes each sequence to the experts matching its generating pattern, comparing against the unspecialized-expert baseline (FAME). Specialization is measured by the CE gap between target and non-target attention distributions (higher is better), and routing quality by routing recovery (higher is better; chance is $1/11 = 0.091$for 11 experts). Results appear in Table~\ref{tab:synth_attn_rout}, with confusion matrices in Appendix~\ref{appendix:synth_interp_viz}. MoTIF exhibits a clear diagonal-versus-off-diagonal CE gap of 2.850$\pm$0.116, confirming specialized attention, whereas FAME shows no such structure (CE$\approx 12$everywhere, gap$\approx 0$), as expected for an unspecialized baseline. Routing recovery for MoTIF is well above chance and near-perfect for cyclical experts, with modest smearing across adjacent fine-range experts, while FAME remains at chance throughout. These results show that MoTIF disentangles underlying temporal patterns in an interpretable, unsupervised manner that is directly recoverable from its routing behavior.

\begin{table}[t]
\caption{\textbf{Temporal attention alignment and routing recovery on synthetic data.} \textit{Top half}: Expert specialization - alignment diag/off-diag CE refers to the matrix of cross-entropy (CE) between target and actual attention distributions for all pairs of experts, where lower CE indicates more agreement. \textit{Bottom half}: Routing recovery - how often the correct expert is assigned the top routing weight.}
\label{tab:synth_attn_rout}
\begin{center}
\resizebox{0.8\textwidth}{!}{
\begin{tabular}{lccc}
\toprule
\bf Metric  & \bf MoTIF (ours) & \bf FAME (no specialization)\\
\hline\noalign{\smallskip}
alignment diag CE (own kernel, $\downarrow$) & \textbf{4.009 ± 0.010} & 12.046 ± 0.470\\
alignment off-diag CE & 6.859 ± 0.106 & 12.072 ± 0.457\\
alignment gap (off-diag minus diag $\uparrow$) & \textbf{2.850 ± 0.116} & 0.026 ± 0.014\\
\hline\noalign{\smallskip}
routing recovery - range mean & \textbf{0.389 ± 0.004} & 0.092 ± 0.022\\
routing recovery - cyclical mean & \textbf{0.829 ± 0.050} & 0.089 ± 0.032\\
\bottomrule
\end{tabular}
}
\end{center}
\end{table}

This interpretability translates into predictive gains: on next-value prediction in the synthetic dataset (Table~\ref{tab:synth_perf}), MoTIF outperforms all baselines across every metric, with the largest gains in F1-macro, MAE, and RMSE, and narrower but consistent gains in accuracy and within-1 accuracy. The narrower margins reflect near-saturation of an analytic oracle ceiling (acc = 0.7462, within-1 = 0.9350, macro-F1 = 0.6903), derived from the known generative functions via a Monte-Carlo estimate of the most frequently sampled bucket over the irreducible sources of uncertainty (time to next event and generative noise). Accuracy and within-1 accuracy approach this ceiling for all methods, but a larger gap remains for macro-F1, where MoTIF comes closest to the oracle. Together, these results confirm that MoTIF's pattern-aligned, label-free routing yields concrete predictive gains, achieving interpretability without sacrificing, and in fact improving, performance whenever the data exhibit cyclical or autocorrelated temporal structure.

\begin{table}[t]
\caption{\textbf{Next value prediction performance comparison on synthetic data.} Continuous metrics (MAE, RMSE) on value bucket index and classification metrics on the next value bucket
(F1 macro, accuracy, within 1). Bold indicates top metric.}
\label{tab:synth_perf}
\begin{center}
\resizebox{\textwidth}{!}{
\begin{tabular}{lccccc}
\toprule
\bf Method  & \bf MAE & \bf RMSE & \bf F1 macro & \bf Accuracy & \bf Within 1\\
\hline\noalign{\smallskip}

DuoRec & 0.3612 $\pm$ 0.0014 & 0.7813 $\pm$ 0.0023 & 0.5552 $\pm$ 0.0190 & 0.7313 $\pm$ 0.0008 & 0.9327 $\pm$ 0.0004\\

SASRec & 0.3574 $\pm$ 0.0025 & 0.7784 $\pm$ 0.0036 & 0.5969 $\pm$ 0.0035 & 0.7332 $\pm$ 0.0015 & 0.9329 $\pm$ 0.0007\\

HSTU & 0.3627 $\pm$ 0.0003 & 0.7835 $\pm$ 0.0010 & 0.5795 $\pm$ 0.0056 & 0.7305 $\pm$ 0.0002 & 0.9322 $\pm$ 0.0002\\

PATHER & {0.3576 $\pm$ 0.0003} & {0.7690 $\pm$ 0.0009} & {0.6421 $\pm$ 0.0023} & {0.7345 $\pm$ 0.0001} & {0.9355 $\pm$ 0.0002}\\

Transformer & 0.3609 $\pm$ 0.0010 & 0.7818 $\pm$ 0.0020 & 0.6108 $\pm$ 0.0082 & 0.7320 $\pm$ 0.0006 & 0.9325 $\pm$ 0.0005\\

FAME & 0.3600 $\pm$ 0.0007 & 0.7811 $\pm$ 0.0013 & 0.6349 $\pm$ 0.0156 & 0.7330 $\pm$ 0.0004 & 0.9323 $\pm$ 0.0002\\

MoTIF (ours) & \textbf{0.3524 $\pm$ 0.0010} & \textbf{0.7650 $\pm$ 0.0003} & \textbf{0.6574 $\pm$ 0.0031} & \textbf{0.7349 $\pm$ 0.0014} & \textbf{0.9360 $\pm$ 0.0004}\\
\bottomrule
\end{tabular}
}
\end{center}
\end{table}

\subsection{Explainability Without Predictive Cost on Credit Card Transactions}

\textbf{MoTIF learns temporal pattern-aligned specialized attention matrices on CCT dataset.}
As on synthetic data, MoTIF achieves clear attention-based specialization on CCT while FAME does not (Appendix Table \ref{tab:cct_attn_rout}): MoTIF shows a much larger diagonal-to-off-diagonal CE gap (1.752 vs. 0.041) and a near-perfect mean own-kernel rank (0.910 vs. 4.730, near-random for FAME). Since ground-truth pattern labels are unavailable here, we instead check routing spread across experts: both models avoid collapsing onto a few experts, but FAME's more uniform spread is uninformative given its lack of specialization, whereas MoTIF's skewed routing likely reflects the true, uneven distribution of temporal patterns in the data.

\begin{figure}[t]
\begin{center}
\includegraphics[width=1\linewidth]{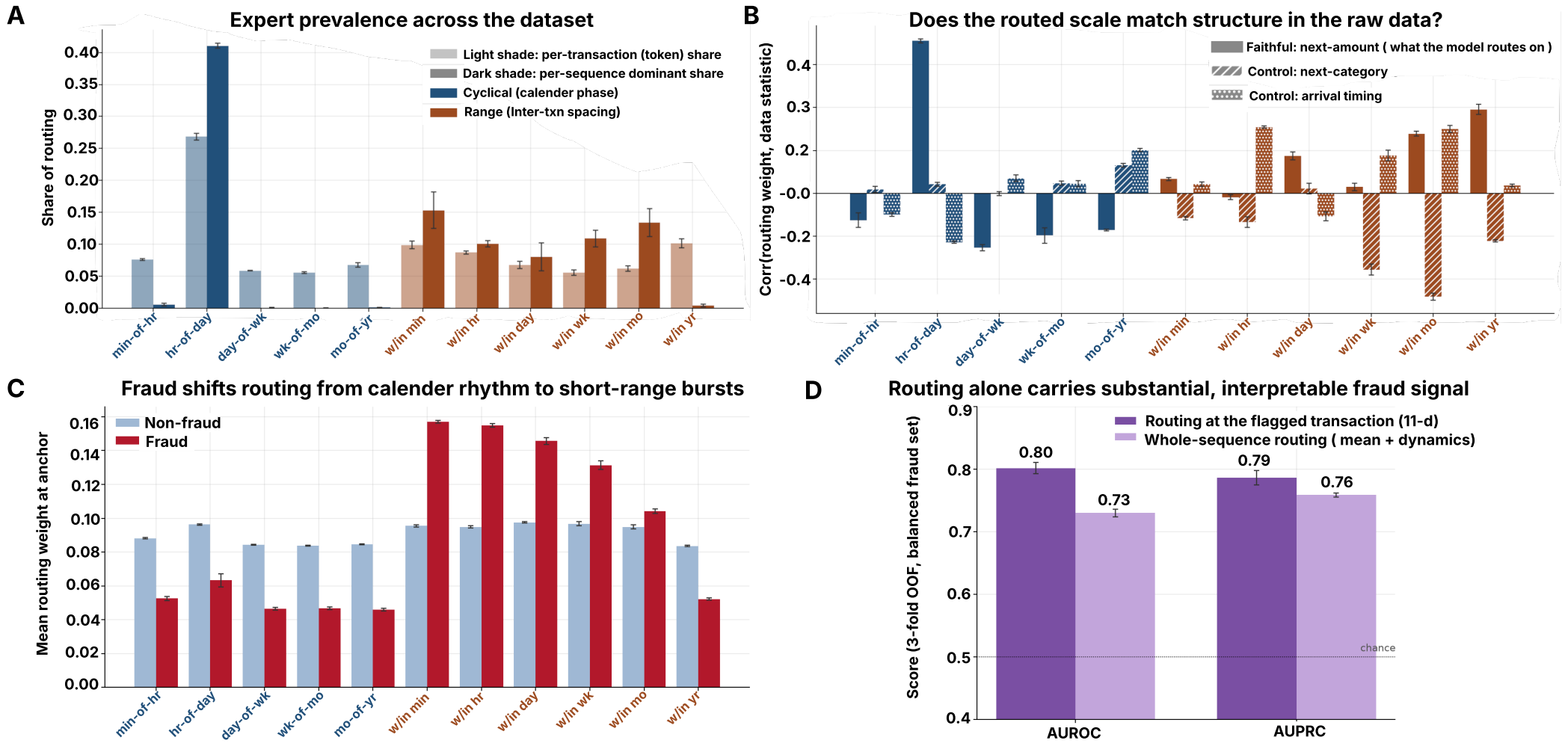}
\end{center}
\caption{\textbf{Routing-based interpretation of temporal patterns in CCT.} (A) Expert prevalence: proportion of top-1 routing per expert, at the transaction level (light) and sequence-dominant level (dark). (B) Validation: correlation between routing weight and raw-data statistics, independent of the router; solid bar reflects next-amount predictability (the model's actual routing signal), hatched bars are controls (next category, arrival timing). (C) Routing by fraud cohort: mean routing weight at the labeled transaction, non-fraud (light blue) vs. fraud (red). (D) Fraud predictability from routing alone: GBT probe performance using transaction-level (dark) vs. sequence-level (light) routing weights; dotted line indicates chance.}
\label{fig:cct_interpretability}
\end{figure}

\textbf{MoTIF interpretability provides insights into credit card transaction patterns.}
MoTIF's routing reveals interpretable temporal structure in CCT (Fig. \ref{fig:cct_interpretability}). Expert prevalence (panel A) shows the cyclical hour-of-day expert dominates at both transaction and sequence level, with sequence-level routing otherwise concentrated among range-based experts (minute-to-month scales), suggesting CCT is largely an intraday-rhythm, short-range dataset. Validating routed sequences against their expert's target pattern for the amount variable (panel B) confirms this: correlations are strongest for the cyclical hour-of-day expert (+0.71), followed by within-year (+0.39), within-month (+0.28), and within-day (+0.18), with a usage-weighted correlation of +0.36; the few negative correlations occur among rarely-used experts with too little signal to fit their pattern. Stratifying by fraud label (panel C) shows non-fraud transactions spread relatively uniformly across experts (calendar-rhythmic legitimate spending), while fraud shifts toward short-range experts and away from cyclical ones, consistent with tight, unscheduled bursts. This routing signal is itself predictive of fraud (panel D: AUROC = 0.80, AUPRC = 0.79), with sequence-level summaries performing comparably but not exceeding transaction-level routing, indicating the fraud signal is concentrated at the transaction itself. Individual-level examples (Appendix Fig. \ref{fig:cct_interpretability_1}) further show per-instance temporal signatures consistent with these cohort-level patterns.

\begin{table}[t]
\centering
\caption{\textbf{Next amount prediction on CCT dataset.} Both continuous metrics (MAE, RMSE) on amount bucket index and classification metrics on the next amount bucket (F1 macro, accuracy, within 1). Bold indicates top metric, underline indicates ours is second best.}
\label{tab:cct_phase1_perf}
\small
\setlength{\tabcolsep}{5pt}
\renewcommand{\arraystretch}{1.08}
\resizebox{\textwidth}{!}{
\begin{tabular}{lccccc}
\toprule
\multicolumn{6}{c}{\textbf{Next Amount Prediction}} \\
\midrule
\textbf{Method} 
& \textbf{MAE} $\downarrow$ 
& \textbf{RMSE} $\downarrow$ 
& \textbf{F1 macro} $\uparrow$ 
& \textbf{Accuracy} $\uparrow$ 
& \textbf{Within 1} $\uparrow$ \\
\midrule
DuoRec      & 0.6863 $\pm$ 0.0019 & 1.3316 $\pm$ 0.0031 & 0.7122 $\pm$ 0.0036 & 0.6208 $\pm$ 0.0006 & 0.8384 $\pm$ 0.0007 \\
SASRec      & 0.9419 $\pm$ 0.0022 & 1.6180 $\pm$ 0.0011 & 0.5565 $\pm$ 0.0024 & 0.5179 $\pm$ 0.0015 & 0.7679 $\pm$ 0.0007 \\
HSTU        & 0.8460 $\pm$ 0.0088 & 1.5073 $\pm$ 0.0104 & 0.6089 $\pm$ 0.0051 & 0.5490 $\pm$ 0.0036 & 0.7972 $\pm$ 0.0023 \\
PANTHER     & 0.7126 $\pm$ 0.0023 & 1.3195 $\pm$ 0.0031 & 0.6270 $\pm$ 0.0026 & 0.5849 $\pm$ 0.0014 & 0.8377 $\pm$ 0.0009 \\
Transformer & 0.8177 $\pm$ 0.0038 & 1.4840 $\pm$ 0.0052 & 0.6638 $\pm$ 0.0004 & 0.5679 $\pm$ 0.0015 & 0.8015 $\pm$ 0.0008 \\
FAME        & {0.5730 $\pm$ 0.0082} & {1.1943 $\pm$ 0.0082} & {0.7320 $\pm$ 0.0031 } & \textbf{0.6655 $\pm$ 0.0051} & {0.8766 $\pm$ 0.0027} \\
\textbf{MoTIF (ours)}
            & \textbf{0.5581 $\pm$ 0.0058}
            & \textbf{1.1484 $\pm$ 0.0127}
            & \textbf{0.7374 $\pm$ 0.0039}
            & \underline{0.6615 $\pm$ 0.0075}
            & \textbf{0.8808 $\pm$ 0.0022} \\
\bottomrule
\end{tabular}
}
\begin{minipage}{0.95\linewidth}
\footnotesize
\end{minipage}
\end{table}

\textbf{MoTIF maintains competitive predictive performance.}
On next amount prediction (Table \ref{tab:cct_phase1_perf}), MoTIF outperforms all baselines except on accuracy, with the largest gains on MAE, RMSE, and F1 macro, again indicating stronger performance on scarcer amount buckets. Next item category prediction shows a slightly different pattern, likely because the routing supervision teacher was trained on amount alone (Appendix~\ref{appendix:cct_itemcat}). On downstream fraud prediction (DCN head over learned embeddings plus handcrafted features, Appendix Table \ref{tab:cct_phase2_perf}), MoTIF and FAME perform notably better than remaining baselines and are comparable to each other, with FAME slightly ahead on average except recall. An embedding-only GBT probe excluding handcrafted features (Appendix Table \ref{tab:cct_phase2_perf_embedonly}) shows MoTIF outperforming all baselines on F1 and AUROC and ranking second on AUPRC and recall, confirming the learned embeddings retain strong predictive value on their own.

Overall, MoTIF matches or exceeds top-performing baselines on next-token prediction and fraud detection on the CCT dataset while providing label-free routing grounded in the data's temporal structure, exposing the specific patterns underlying individual predictions.

\section{Conclusions}
We introduce MoTIF, an interpretable framework that decomposes irregular time sequences into temporal patterns across scales (minute through year) via an attention query mixture-of-experts, using temporal alignment and a novel label-free routing supervision mechanism requiring no ground-truth labels. On synthetic data, MoTIF recovers correct expert specialization and routing while improving next-token prediction, showing interpretability and predictive gains are jointly achievable. On a public credit card dataset, MoTIF preserves prediction performance and fraud-detection generalizability while revealing interpretable structure: intraday-rhythm and short-range patterns dominate overall, and fraudulent transactions cluster in short bursts with no calendar regularity, unlike the regular rhythms of non-fraudulent ones. These results show MoTIF exposes dataset-wide, cohort-level, and individual behavior without sacrificing performance.

Limitations suggest future directions. Our routing teacher anchors to a single feature (next amount), weakening routing for non-anchor features under multitask prediction; task-specific multi-signal routing is a natural extension. We model patterns in the value signal rather than timestamps directly, so explicitly modeling inter-event gaps could better capture burstiness. Our enforced temporal trends may also not capture more complex or irregular structure, though MoTIF's bandwidth parameter offers some flexibility. Overall, MoTIF shows interpretable structure can reveal the temporal drivers of behavior and outcomes like fraud without sacrificing predictive performance, pointing toward more transparent sequence models for high-stakes settings.

\section{AI Use Statement}
In this work, we used generative AI tools to help formulate mathematical claims, refine hypotheses, provide feedback on research methodology, design experiments, implement methods, clean and reformat datasets, support qualitative and thematic data analysis, and interpret results. We have not used generative AI tools for generating synthetic datasets, developing theoretical models or conceptual frameworks, providing critical ingredients for proving mathematical claims, proposing hypotheses, designing the methodology, or writing proofs, and translation is not applicable to this work. Additionally, we used generative AI tools to help create scientific figures, suggest experimental parameters, identify and summarize existing related literature, and edit parts of the paper to improve readability. We have reviewed all AI-assisted work.  LLM-generated code was validated via unit tests and visualizations. Conceptual insights were treated as supplementary viewpoints and critically evaluated before inclusion. All citation and reference suggestions were manually checked for accuracy. We take responsibility for the final content of this work, including text, claims or artifacts produced with the aid of generative AI.

\section{Ethics Statement}
The data used in this work are either synthetically generated or derived from a publicly available credit card transaction dataset comprising synthetic consumers generated via a multi-agent virtual world simulation by IBM (\cite{altman2021synthesizing}). Consequently, these experiments involve no real human subjects, eliminating risks associated with the exposure of personally identifiable information (PII). While the synthetic data mirror broad U.S. demographic distributions (e.g., age, income, geographic distribution), fraud labels were generated based on real observations on fraudster behavior (\cite{altman2021synthesizing}). However, exploratory analyses revealed demographic variations in baseline features (such as a higher prevalence of fraud in specific older age cohorts; see Appendix Fig. \ref{tab:cct_demographics}). We acknowledge that synthetic datasets can inadvertently capture or amplify systemic correlations. We encourage future practitioners deploying or building upon our model to conduct rigorous fairness audits and implement mitigation strategies tailored to their specific deployment context to prevent the perpetuation of downstream harms.

\section{Reproducibility Statement}
To facilitate the reproducibility of our work, we supply a link to an anonymous repository containing the source code used to run the experiments in this paper: \url{https://anonymous.4open.science/r/MoTIF-82E4}. The methodology, training and implementation details, and evaluation metrics are described in the Method section and Appendix. We provide a reference containing the link to the public credit card transactions dataset used in this paper, along with details on preprocessing within the Appendix. For the synthetic dataset, we specify how these data were generated in the Appendix and provide related code in the anonymous repository.

\bibliography{iclr2025_conference}
\bibliographystyle{iclr2025_conference}

\newpage
\appendix
\section*{\centerline{Appendix}}
This appendix provides additional details and results for ``Decomposing Irregular Time Series into Interpretable Temporal Patterns with Mixture-of-Experts". Preliminary background on temporal pattern decomposition is presented in Section~\ref{appendix:preliminaries}. Details on the synthetic dataset generation and credit card transaction dataset preprocessing are provided in Section~\ref{appendix:dataset_details}. Additional methodological, training and evaluation details can be found in Section~\ref{appendix:addl_implementation_details}. Further results demonstrating temporal pattern specialization and prediction performance are shown for the synthetic and credit card transaction datasets in Sections~\ref{appendix:addl_synth_results} and \ref{appendix:addl_cct_results}, respectively.

\section{Preliminaries}\label{appendix:preliminaries}
\textbf{Definition: Temporal Pattern Decomposition.} We define temporal pattern decomposition as assigning a time-indexed sequence to its constituent cyclical and elapsed-range patterns at the minute, hour, day, week, month, and year scales. Assignment weights quantify both the predictive utility of each pattern and its agreement with the autocorrelation structure observed in the data.

\textbf{Problem Statement.} Given a set of irregularly spaced time sequences $X = \{\mathbf{x_s}\}_{s=1}^N$, where $\mathbf{x_s} = [x_1,...,x_{T_s}]$ is a length-$T_s$ sequence of $d$-dimensional feature vectors $x_i \in \mathbb{R}^d$, we aim to identify each sequence's characteristic cyclical and elapsed-range patterns at time-point granularity. We learn a transformer with a mixture-of-experts (MoE)-based self-attention query, where each expert $e$ specializes to a particular temporal pattern (cyclical or autocorrelative) and scale (minute, hour, day, etc.) via its induced attention matrix. A routing function $\mathcal{R}$ produces weights $\pi_{s,i}^{(e)}$ quantifying each pattern's relevance to token $x_{s,i}$, determining each expert's contribution to the final query and, in turn, the strength of its attention pattern at each time point. The output of $\mathcal{R}$ thus identifies the predominant temporal patterns within $\mathbf{x_s}$.

\section{Dataset Details}\label{appendix:dataset_details}
\subsection{Synthetic Dataset Generation}\label{appendix:synthetic}
To demonstrate that our label-free routing supervision can recover known temporal patterns in the data without being explicitly trained on the known temporal pattern labels, we conduct a proof-of-concept study on synthetic irregular time sequence data containing known temporal patterns. The synthetic dataset consists of sequences, each of which is generated according to one of 17 temporal patterns matched to those modeled by our 11 experts. The patterns break down into: (a) five cyclical: minute-of-hour, hour-of-day, day-of-week, week-of-month,  month-of-year; and (b) 12 range-based, comprising six main variants (within minute, hour, day, week, month, year), with two observation variants each, linear and nonlinear. For each sequence, a set of time stamps was generated, followed by the application of the corresponding pattern-specific function to those time stamps to generate the signal variable (analogous to ``amount" in transaction sequence data) at each sampled time point. We generated 300 train, 64 validation and 64 test examples per pattern, for a total of 5,100 training, 1,088 validation, and 1,088 test samples. All synthetic sequences have length $L=1024$.

\textbf{Timestamp generation.} Timestamps were generated from the same underlying distribution across all temporal patterns. This ensured the model would learn to route to specialized experts based on the `amount` pattern signal, rather than based on timing statistics. All sequences draw
$L-1$ inter-transaction gaps from the log-uniform distribution over $[1 \text{ second}, 1 \text{ month}]$, i.e., in hours: $\Delta t_i \sim \exp(U(\ln(1/3600), \ln(630)))$, such that $\boldsymbol{\Delta} = [0, \Delta t_1, ..., \Delta t_{L-1}]$ and $t_i = \sum_{k \leq i} \Delta_k$. This enables us to cover sub-minute to month gaps, while spanning multiple years to enable within-year range windows and month-of-year cyclical patterns. Drawing the gaps from the log uniform distribution also ensures that patterns on all scales are evenly resolvable; thus, timestamp sampling can be shared across patterns so that temporal density cannot be a shortcut for the router. For each of the below generating patterns, we empirically hand-tuned the hyperparameters to maintain a good signal-to-noise ratio and realistic-looking sequences.

\textbf{Cyclical pattern generation.} For cyclical patterns, we model the amount using a truncated Fourier series of the calendar phase $F_i$ (e.g. 1-24 for hour-of-day; 1-7 for day-of-week), with period $P$ in units of the scale of interest (e.g. for hour-of-day, $P = 24$; for day-of-week, $P=7$). The generated time sequence is given by:
\begin{align}
    \theta_i &= 2\pi \cdot \frac{F_i \bmod P}{P}\\
    x_i &= \mu+A\sum_{m=1}^H a_m \cos(m\cdot \theta_i + \varphi_m)+\epsilon_i, \quad \epsilon \sim N(0,\sigma^2)
\end{align}
where harmonic $H=2$; $\mu \sim N(3.33, 0.8)$ clipped to [1.0, 6.5] representing a sequence's average value (e.g. average log amount spent on a card); coefficients $a_m \sim U(0.4,1.0)\cdot \pm 1$ and normalized so $\sum |a_m| = 1$; amplitude $A \sim U(0.4, 1.3)$; phases $\varphi_m \sim U(0,2\pi)$; and noise $\sigma \sim U(0.03, 0.10)$. Using $H=2$ enables a more complex signal than a pure sinusoid while maintaining the clear periodic pattern and smoothness. Because amount is phase-locked, transactions occurring at the same phase will have similar amounts regardless of the amount of time that passed between them, mirroring cyclical spending habits.

\textbf{Range-based pattern generation.} We modeled range-based patterns as autoregressive functions in which the value at a given time depends only on the values within the preceding range, i.e., each pattern is distinguished by a characteristic correlation timescale. This is parameterized using the Ornstein-Uhlenbeck (OU) process, which directly models signal decorrelation over time while handling irregular timestamps via continuous-time functions that depend on time elapsed, rather than timestep index. Each range expert corresponds to a specific time scale $\tau$ in hours: 1/60 (minute), 1 (hour), 24 (day), 168 (week), 730.5 (month), and 8766 (year). To handle irregular time sampling, the latent OU state depends on the actual timestamps:
\begin{align}
    &a_i = \exp(-\Delta_i / \tau)\\
    &dX_0 = \sigma z_0\\
    &dX_i = a_i dX_{i-1} + \sigma\sqrt{1-a_i^2} \cdot z_i, \quad z_i \overset{\text{iid}}{\sim} N(0,1)
\end{align}
where $\Delta_i$ is the elapsed hours between transaction $i-1$ and $i$; $a_i \in (0,1]$ is the memory factor controlling how much of the previous deviation survives across the time gap; $dX_i = x_i - \mu$ is the deviation from baseline (OU ``state"); $\sigma \sim U(0.5, 0.9)$ is the stationary standard deviation, fixed for each sequence; and $z_i$ is the innovation, i.e. the new information that was not predictable from the previous state. We generate the time sequence values based on two variants: linear and nonlinear. For the linear variant:
\begin{equation}
    x_i = \mu+dX_i
\end{equation}
where $x_i$ is the observed value at transaction $i$ (e.g. log amount) and $\mu$ is the per-sequence baseline that the process reverts to, i.e., long-run mean. For the nonlinear variant, we apply a nonlinear warp $W(\cdot)$:
\begin{equation}
    x_i = \mu + W(dX_i), \quad  W(v) = v + \sum_{m=1}^M \alpha_m \sin(\omega_m v+\psi_m)
\end{equation}
for $M=3$; amplitude $\alpha \sim U(0.4,1.0) \cdot \pm 1$ normalized and scaled by $a \in U(0.6,1.0)$; angular frequency $\omega \sim U(1.5,3.0)$; and phase offset $\psi \in U(0,2\pi)$.

We note that because time sampling density and signal variance are held constant across patterns, the model cannot rely on these for routing. Instead, the model must understand the temporal structure of the amount signal to route a sequence to the correct expert. For cyclical, this is phase-locking to a calendar field, whereas for range, this is the OU autocorrelation decay length $\tau$.

\subsection{Credit Card Transasction (CCT) Dataset}\label{appendix:cct}
We apply our method to credit card transaction (CCT) data from Kaggle (\cite{cct}). This dataset was generated from a multi-agent virtual world simulation performed by IBM, containing 4,731,394 transactions from 5,777 accounts spanning decades of purchases. We evaluate the performance of our model on both autoregressive next amount bucket and next item category tasks (phase 1) and downstream fraud detection (phase 2). Phase 1 is the pretraining phase, learning to generate representations that capture the temporal patterns underlying the sequence that are predictive of future transactions. Phase 2 allows us to evaluate how well our representations can generalize to a new task. Features include both per-transaction (merchant category, hour of day, purchase amount, number of hours since previous transaction, and transaction direction, and payment method) and user profile-level information (FICO score, credit limit, age, gender, income, number of credit cards, and card information). 

For the phase 1 general pretraining step, we subset to the non-fraud samples with sequence length at least 30. We then conduct a nested temporal train/val/test split within each sequence as follows: first, truncate each sequence to the last $N=1024$ transactions, then set the train split to the first $N-20$ transactions, the validation set to the first $N-10$ transactions, and the test split to the full set of transactions. In this way, the validation and test set each contain an additional 10 held-out next-token targets. The features included in phase 1 training are the transaction-level sequence features
and the account profile features.

For phase 2, we perform fraud detection at the transaction-level using DCN and GBT probes for transaction-time and next-transaction fraud scoring, respectively. We first select the transactions whose labels will be predicted. To handle class imbalance, we select all fraud transactions, and then undersample the non-fraud transactions to the same sample size as the fraud transactions. We split the transactions of interest into 68\% train, 17\% validation, and 15\% fraud. For transaction-time fraud detection, the input data used to predict the fraud label include hand-engineered fraud features (including rolling-30-day amount stats, ratios, and account attributes) along with the 1024-transaction window ending at and including that transaction. We designed this splitting and training mechanism to match that of the PANTHER (\cite{li2026panther}) baseline. For the next-transaction fraud scoring, we only input the transaction history embedding to assess embedding quality against the baseline methods.

\subsection{Credit Card Transactions Demographics by Fraud Label}

\begin{table}[th!]
\caption{\textbf{Demographic features by fraud label.} Reported values are mean $\pm$ standard deviation for age, income, debt, and FICO score, separated by fraud label. For gender, we report female/male counts per fraud label. Individuals are labeled as fraudsters if they have at least one fraudulent transaction.}
\label{tab:cct_demographics}
\begin{center}
\resizebox{0.8\textwidth}{!}{
\begin{tabular}{lcc}
\toprule
\bf Feature  & \bf Non-Fraud & \bf Fraud\\
\hline\noalign{\smallskip}
Age & 30.57 $\pm$ 14.44 & 52.64 $\pm$ 15.59 \\
Gender (female/male) & 329/328 & 687/656 \\
Per Capita Income - By Zipcode & \$23,084.63 $\pm$ \$10,804.45 & \$23,169.96 $\pm$ \$11,573.71 \\
Yearly Income - By Person & \$46,611.70 $\pm$ \$22,003.43 & \$45,277.64 $\pm$ \$23,456.91\\
Total Debt & \$73,498.22 $\pm$ \$50,976.28 & \$58,921.12 $\pm$ \$52,222.65\\
FICO Score & 702.90 $\pm$ 69.33 & 713.08 $\pm$ 65.94 \\
\bottomrule
\end{tabular}
}
\end{center}
\end{table}

\section{Additional Implementation Details}\label{appendix:addl_implementation_details}
\subsection{Multi-head MoE-based query attention}\label{appendix:multihead_query_attn}
This is implemented within a \textit{multi-head} framework, using a set of projectors with output dimension $d'=d/H$ to split the embedding dimension $d$ into $H$ heads. The result of the MoE component is a set of query vectors $q_i$ for each time point, $i$, for each head $h$, resulting in $Q_h = [q_1, q_2, ..., q_T]$. Each query vector is a mixture of the expert-specific queries, weighted by the router: given the set of query vectors $q_i^e$ output by each expert $e$, the final query $q_i$ is a gated combination of the expert queries:
\begin{equation}
    q_i = \sum_{e \in E} \pi_{i,e}q_i^e
\end{equation}
where $\pi_{i,e} \geq 0$ and $\sum_{e\in E} \pi_{i,e} =1$. Key and value vectors are obtained using the weight matrices $W_h^K$ and $W_h^V$, as $K_h = XW_h^K$ and $V_h = XW_h^V$. Attention heads are computed as
\begin{equation}
\text{head}_h = \text{Attention}(Q_h,K_h,V_h) = \text{softmax}\left(\frac{Q_hK_h^\top}{\sqrt{d'}}\right)V_h
\end{equation}
The final representations ${z_1},{z_2},...,{z_T}$ are obtained as:
\begin{equation}
\text{MultiHead}(Q,K,V) = \text{concat}(\text{head}_1,\text{head}_2,...,\text{head}_h)W^O  
\end{equation}
where $W^O$ is a final weight matrix. This produces a set of embeddings for each position in the sequence. These are aggregated into a single sequence-level embedding via attention pooling. This uses a learnable weight vector ${w_{pool}}$ that projects the input embeddings to a scalar score, $c_i$. These scores are normalized via softmax to sum to 1, such that $b_i = \text{softmax}(c_i)$, and then the final sequence-level embedding is the weighted sum across position embeddings:
\begin{equation}
    {z_{seq}} = \sum_{i=1}^T b_i \cdot {z_i}, \quad\quad \text{where} \quad\quad b_i = \frac{e^{c_i}}{\sum_{j=1}^T e^{c_j}} \quad\quad \text{and} \quad\quad c_i = {w_{pool}}^\top {z_i}
\end{equation}

\subsection{Training Objectives}\label{appendix:training_details}
\textbf{Phase 1: Pretraining on Next Amount and Category Prediction}
Next-token prediction uses per-task linear heads (e.g., next amount bucket, next item category), each scored via cross-entropy over target positions:
\begin{equation}
    \mathcal{L}_{\text{task}} = -\frac{1}{N}\sum_{i}^N \sum_{c \in C} y_{i,c} \log\hat{y}_{i,c}
\end{equation}
where $\hat{y}_{i,c}$ is the predicted logit for class $c$ and example $i$, and $y_{i,c}$ is the one-hot true class. Pretraining on these tasks teaches the model to extract temporal patterns from the data. The full pretraining objective combines the task loss with attention expert specialization ($\mathcal{L}_{\text{align}}$), temporal pattern-aware routing ($\mathcal{L}_{\text{route}}$), a standard routing load-balance loss \citep{shazeer2017outrageously} to prevent collapse onto a single expert ($\mathcal{L}_{\text{lb}}$), and a sparse top-$k$ load-balance loss with $k=1$ to prevent the same experts from consistently dominating ($\mathcal{L}_{\text{lbtk}}$):
\begin{equation}
    \mathcal{L}_{\text{phase}_1} = \sum_{i} \mathcal{L}_{\text{task}_i} + \lambda_{\text{align}}\mathcal{L}_{\text{align}} + \lambda_{\text{route}}\mathcal{L}_{\text{route}} + \lambda_{\text{lb}}\mathcal{L}_{\text{lb}} + \lambda_{\text{lbtk}}\mathcal{L}_{\text{lbtk}}
\end{equation}
The $\lambda$ terms are scalar weights that control the influence of each component of the objective function. These were fine-tuned on the validation set. For the CCT dataset, $\lambda_{\text{align}} = 0.05$, $\lambda_{\text{route}} = 0.3$, $\lambda_{\text{lb}} = 0.01$, and $\lambda_{\text{lbtk}} = 0.1$. For the synthetic dataset, $\lambda_{\text{align}} = 0.1$, $\lambda_{\text{route}} = 1.0$, $\lambda_{\text{lb}} = 0.01$, and $\lambda_{\text{lbtk}} = 0.1$.

\textbf{Phase 2: Downstream Fraud Prediction}
To evaluate how well the pretrained sequence embeddings generalize, we fine-tune on fraud prediction by replacing the next-token heads with a binary fraud head, freezing the backbone, and training only the fraud head. The fraud head is a deep and cross network (DCN) \citep{wang2017deep}, chosen for its explicit modeling of feature interactions. It takes as input the frozen model's sequence embedding along with hand-engineered features (sequence-level statistics and account profile features), predicting fraud at target transactions via binary cross-entropy:
\begin{equation}
    \mathcal{L}_{\text{fraud}} = -\frac{1}{|{A}|}\sum_{i\in A} \Big[y_{i}\log\sigma(\hat{y}_{i}) +
  (1-y_{i})\log\big(1-\sigma(\hat{y}_{i})\big)\Big]
\end{equation}
where $A$ is the set of target transactions, $\hat{y}$ the predicted logit, $y \in \{0,1\}$ the true fraud label, and $\sigma$ the sigmoid function. To isolate the fraud-discriminative signal in the frozen embeddings alone, we additionally train an independent gradient boosted tree (GBT) probe \citep{friedman2001greedy} on the embeddings without auxiliary features or the DCN head.

\subsection{Training Details}\label{appendix:impl_details}
For the experiments herein, we fixed the model architecture with embedding dimension 512, 2 heads, and 2 layers. We set dropout to 0.1 and max sequence length to 1024. The MoE layer comprises 11 total experts (5 cyclical and 6 range). On the synthetic data, we used route weight 1.0, temperature 0.5, cyclical-first gate threshold of 0.90, $w_{fam} = 1.0$, $w_{cyc} = 1.0$, $w_{rng} = 2.0$, temporal alignment weight 0.1, top-k load balance weight 0.1 for $k=1$, and soft routing auxiliary loss 0.01. For the CCT dataset, route weight was 0.3, $w_{rng} = 1.0$, and temporal alignment weight was 0.05; the rest was the same as on the synthetic data. For training, we used a batch size of 128,  AdamW optimizer with weight decay of 0.01, cosine learning rate schedule with 500 warmup steps and minimum learning rate ratio 0.1. On the synthetic data, peak learning rate was $3\times10^{-4}$, with 5K training steps. For the CCT dataset, we used peak learning rate $6\times 10^{-4}$ and 25K training steps. Models were trained on the training set and parameters were tuned on the validation set. We trained each model using three seeds: 42, 43, 44. Reported results are mean $\pm$ std over these seeds on a held-out test set. For phase 2 on the CCT dataset, we used a frozen backbone and added a DCN head, trained for 4900 steps with learning rate $1\times 10^{-3}$, batch size 256, and 2 cross layers. For the GBT embedding-only probe, we used 150 max iterations, learning rate 0.08, and early stopping, with max leaf nodes set to 31 and max bins to 255.

Model training and experiments were carried out on a single NVIDIA B200 GPU per run, using CUDA 12.9. Peak GPU memory for our MoTIF was $\sim 157$GB. We used Python 3.12.10, PyTorch 2.9.1+cu128, Hydra 1.3.2. Phase 1 training on synthetic data at 5K steps took $\sim$1.46h (0.95 steps/s); on CCT data at 25K steps, training took $\sim$ 5.2 h (1.44 steps/s). Phase 2 CCT fraud prediction training at 2K steps took $\sim$8 min.

\subsection{Evaluation Metrics}\label{appendix:eval_metrics}
We evaluate performance on both interpretability, via attention specialization and routing, and prediction performance. For interpretability, we assess expert-specific attention matrix specialization via alignment cross-entropy (CE) between experts $e_1$ and $e_2$, defined as:
\begin{equation}
    \text{Align-CE}_{e_1,e_2} = -\sum_{i,j} t^{(e_1)}_{i,j}\log 
  a^{(e_2)}_{i,j}
\end{equation}
where $t^{(e_1)}_{i,j} = W^{e_1}_{i,j}\cdot \mathbf{1}[j<i]$ is the expert $e_1$'s causal-masked target attention kernel and $a^{(e_2)}_{i,j}$ is expert $e_2$'s learned attention between time points $i$ and $j$. This measures agreement between the learned and target attention matrices, where lower indicates a better match. We measure the mean diagonal (diag), where $e_1 = e_2$, and off-diagonal (off), where $e_1 \neq e_2$, values of the $\text{Align-CE}$ matrix and compute their gap = off - diag. Larger positive gap indicates better differentiation between the attention matrices of different experts toward their target values.

We assess routing on the synthetic data via routing recovery. This is only applicable when true temporal pattern labels are known. Taking the top-routed expert to be that with the largest router weight, \textit{per-pattern} recovery is given by 
\begin{equation}
    \text{RoutingRecovery}_e = \frac{C_{i,e}}{\sum_j{C_{i,j}}}
\end{equation}
where $i$ is the true pattern, the expert assigned to that pattern is $e$, and $C_{i,j}$ is the count confusion matrix between pattern $i$ and top-routed expert $j$. Thus, this value is the proportion of sequences with true pattern $i$ for which the top-routed expert was the expert specialized for that pattern (expert $e$). Chance is 1/(num experts) = 0.091.

On the CCT dataset which does not have known temporal pattern labels, we assess routing health via entropy ratio (HRatio):
\begin{equation}
    \text{HRatio} = -\frac{\sum_i^{|E|} p_i \ln p_i}{\ln N}
\end{equation}
where $E$ is the set of all experts. For count vector $c$ over $|E|=11$ experts, $p_i = c_i/\sum_j^{|E|}c_j$. At the token level (`token-dispatch'), $c_i$ represents the number of tokens whose top-1 expert is $i$. At the sequence-level (`across-seq'), $c_i$ is the number of sequences whose dominant expert (expert with the largest sequence mean routing) is $i$. Higher ratio (max 1) indicates the top-1 expert is more evenly spread across all possible choices; low ratio (min 0) points towards collapse onto few experts.

We assess concordance between the top-routed expert and temporal pattern detected from the data using correlation between the routing weight and a statistic measuring the strength of that pattern within the data. For next amount prediction, we measure the strength of expert $e$'s temporal pattern in the data via point-forecast $R^2$ using the target attention kernel $W^e$ corresponding to that expert:
\begin{align}
    &R^2_{s,e}=1-\frac{\text{mean}_i(\text{pred}_{e,i}-x_i)^2}{\text{Var}_s(x)}\\
    &\text{pred}_{e,i}=\frac{\sum_{j<i}W^e_{ij}x_j}{\sum_{j<i}W^e_{ij}}
\end{align}
where $x_i$ is the sequence value (e.g. amount) at time $i$ and $\text{Var}_s(x)$ is the variance of $x$ within the sequence. Concordance is then measured via Pearson's correlation between the mean routing weight ($\bar{w}_e$) for expert $e$ and the point-forecast $R^2$ of the sequence value using that expert's target kernel:
\begin{equation}
    r_e = \text{corr}_s(\bar{w}_e,R^2_{s,e})
\end{equation}
We also compute usage-weighted correlation as the weighted average across all experts based on the proportion of times each expert is the dominant pattern in a sequence. This provides a measure of overall agreement between top-routed expert and detected patterns while accounting for how much each expert is actually used.

For the next-category control, we replace the statistic $R^2_{s,e}$ with log-loss skill ($\text{skill}_e = 1-\sum_i CE(p_{e,i},c_i)/\sum_i CE(q_i,c_i)$, where $p_{e,i}$ is the predicted probability of category $c_i$ over past time points $j < i$ based on occurrence indicators of category $c_i$ weighted by the expert $e$'s temporal pattern kernel, $q_i$ is the proportion of times $j < i$ where category class $c_i$ appeared and $CE(a,b)$ is the cross-entropy loss given probability $a$ and true class $b$. For the arrival timing control, we replace the statistic with arrival-timing concentration.

For next-amount bucket prediction, we report accuracy, F1 macro (mean of F1 over classes equally weighted), and within 1 (proportion of samples with predicted amount bucket within 1 of the true amount bucket). We also report MAE and RMSE on the bucket indices. For next category prediction (CCT dataset only), we report accuracy and F1 weighted (mean of F1 weighted by class prevalence) to account for strong class imbalance. For downstream fraud prediction, we report AUROC, AUPRC, and precision@$\tau^*$, recall@$\tau^*$, and F1@$\tau^*$ where $\tau^*$ is a threshold optimally tuned on the validation set.

\newpage

\section{Additional Results on Synthetic Data}\label{appendix:addl_synth_results}
\subsection{Synthetic Data Attention Specialization and Routing Recovery Visualizations}\label{appendix:synth_interp_viz}
\begin{figure}[th]
\begin{center}
\includegraphics[width=0.75\linewidth]{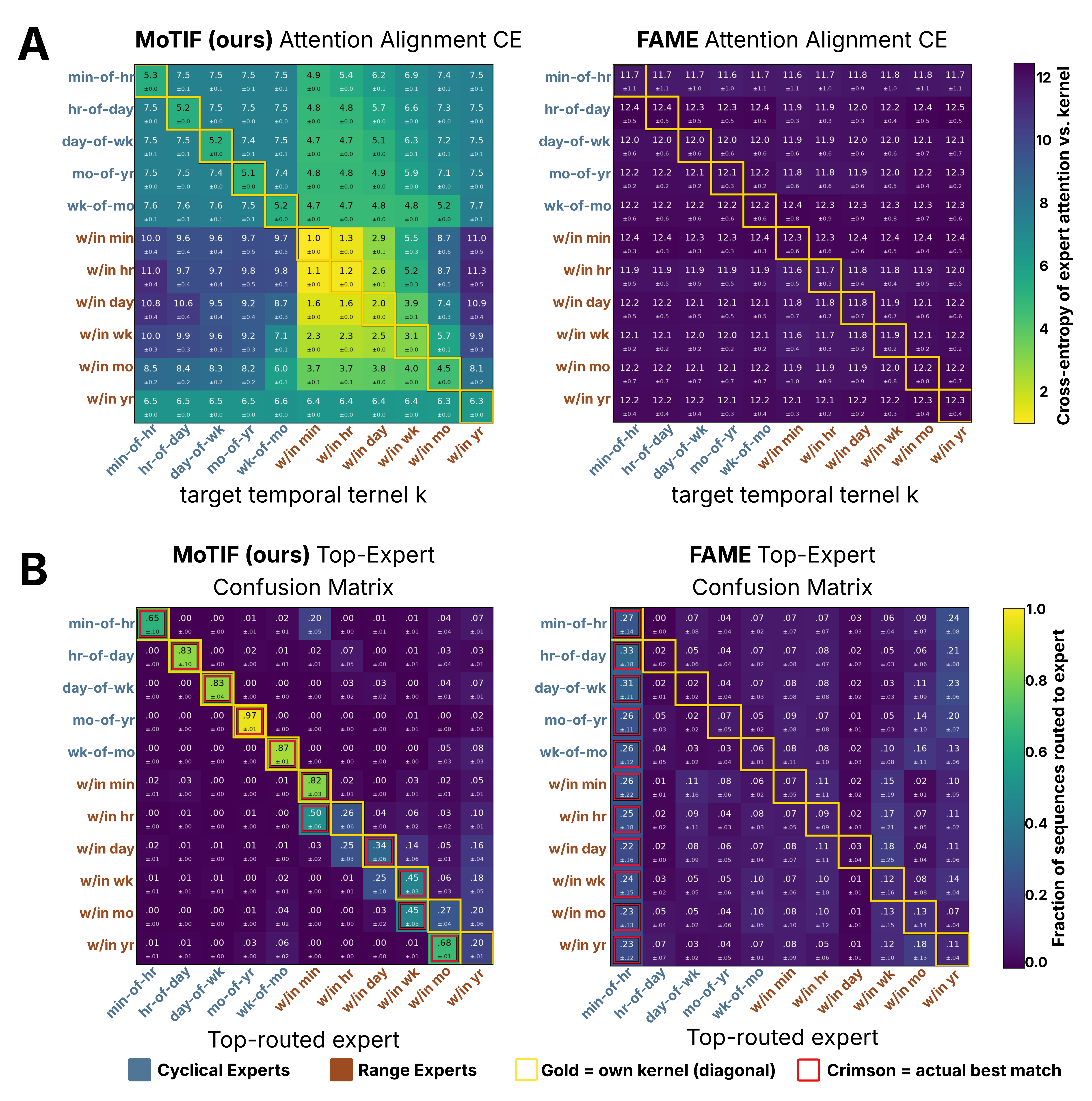}
\end{center}
\caption{\textbf{Attention specialization cross-CE and routing confusion matrices on synthetic data.} Comparing MoTIF (ours, \textit{left}) against FAME (unspecialized attention experts, \textit{right}) (A) Layer-1 expert attention versus target temporal kernels cross-entropy, seed-averaged; gold indicates own kernel (diagonal); lower CE indicates closer match. (B) Layer-1 routing confusion matrix: which expert wins each generating pattern, row-normalized top-expert fractions, seed-averaged; gold indicates the generating pattern's own expert (diagonal); crimson indicates the most-routed expert.}
\label{fig:synth_attn_rout}
\end{figure}

When we visualize the attention alignment cross-entropy (Fig. \ref{fig:synth_attn_rout}A), we can see that cross-entropy tends to be lower between the learned attention matrices and their target kernels, as indicated by lighter colors along the diagonal. This matches our alignment diagonal CE metric being lower than the alignment off-diagonal CE, with a gap of 2.850 $\pm$ 0.116. These results suggest that attention matrix specialization works. On the other hand, the FAME attention CE shows no diagonal versus off-diagonal CE patterns, with values approximately 12 everywhere and the diagonal versus off-diagonal gap of approximately 0, suggesting no specialization, as expected. Additionally, the routing confusion matrices in Fig. \ref{fig:synth_attn_rout}B demonstrate the top routed experts for each known sequence pattern. For our MoTIF method, we see a strong diagonal, demonstrating that the model was able to route patterns to the correct corresponding experts. This shows that the model was able to resolve the temporal patterns via routing to the specialized experts even without being directly supervised on the known pattern labels. Routing was near-perfect for the cyclical experts, with slight smearing among adjacent fine-range experts, which is expected. The FAME baseline, on the other hand, again showed no such specialization, with sequences being most frequently routed to one of the experts, regardless of their underlying temporal pattern. These results demonstrate that MoTIF can effectively untangle the underlying temporal patterns in an interpretable manner that can be read out by observing the routing patterns to the temporally aligned experts.

\section{Additional Results on Credit Card Transactions Data}\label{appendix:addl_cct_results}

\subsection{Expert Specialization and Routing Diversity on Credit Card Transactions}\label{appendix:cct_attn_rout}
\begin{table}[th]
\caption{\textbf{Temporal attention alignment and routing diversity on CCT dataset.} \textit{Top half}: Expert specialization. Alignment diag/off-diag CE refers to the matrix of cross-entropy (CE) between target and actual attention distributions for all pairs of experts - lower CE indicates more agreement. \textit{Bottom half}: Routing diversity versus collapse at both token (token-dispatch) and sequence (across-seq) levels. Higher entropy ratio indicates more uniform routing over experts. Across-seq top-expert share measures how often the primary routed expert is selected.}
\label{tab:cct_attn_rout}
\begin{center}
\resizebox{0.9\textwidth}{!}{
\begin{tabular}{lccc}
\toprule
\bf Metric  & \bf MoTIF (ours) & \bf FAME (no specialization)\\
\hline\noalign{\smallskip}
alignment diag CE (own kernel, $\downarrow$) & \textbf{5.105 ± 0.009} & 8.756 ± 0.004\\
alignment off-diag CE & 6.857 ± 0.012 & 8.798 ± 0.091\\
alignment gap (off-diag minus diag $\uparrow$) & \textbf{1.752 ± 0.007} & 0.041 ± .095\\
\# experts whose own kernel is best (/11) & \textbf{5.330 ± 0.470} & 1.670 ± 0.470\\
mean own-kernel rank (0=perfect, 5=random) & \textbf{0.910 ± 0.070} & 4.730 ± 0.790\\
\hline\noalign{\smallskip}
token-dispatch entropy ratio $\uparrow$ & 0.954 ± 0.003 & 0.991 ± 0.005\\
token-dispatch effective \#experts (/11) & 9.870 ± 0.050 & 10.730 ± 0.120\\
across-seq dominant-expert entropy ratio $\uparrow$ & 0.695 ± 0.006 & 0.931 ± 0.032 \\
across-seq effective \#dominant experts (/11) & 5.300 ± 0.080 & 9.370 ± 0.730 \\
across-seq top-expert share & 0.403 ± 0.006 & 0.177 ± 0.036 \\
\# experts ever dominant (/11) & 10.670 ± 0.470 & 11.000 ± 0.000 \\
\bottomrule
\end{tabular}
}
\end{center}
\end{table}

\subsection{Individual Routing Patterns on Credit Card Transactions}
\begin{figure}[h]
\begin{center}
\includegraphics[width=\linewidth]{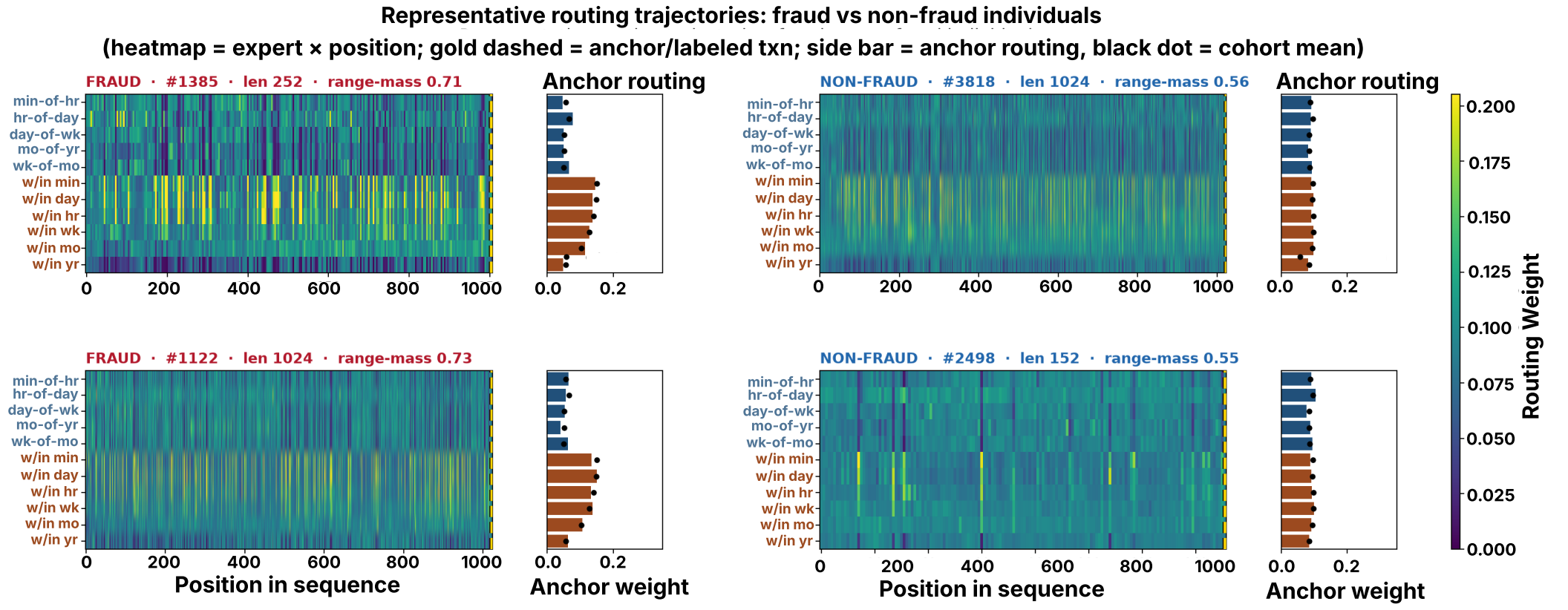}
\end{center}
\caption{\textbf{Individual sequence routing patterns.} Heatmaps indicating routing weight for each expert across the transaction sequence for fraud (left) and non-fraud (right). Gold dashed line marks the labeled transaction (anchor); bar plots show routing weights at the anchor, with black dots indicating cohort mean. Range-mass is the sum of routing weights over the six range experts.}
\label{fig:cct_interpretability_1}
\end{figure}

\newpage

\subsection{Next Item Category Prediction Results on Credit Card Transactions}\label{appendix:cct_itemcat}
\begin{table}[th]
\centering
\caption{\textbf{Next item category prediction on CCT dataset.} Non-ordinal classification only. Bold indicates top metric, underline indicates ours is second best.}
\label{tab:cct_phase1_perf}
\small
\setlength{\tabcolsep}{5pt}
\renewcommand{\arraystretch}{1.08}

\begin{tabular}{lcc}
\toprule
\multicolumn{3}{c}{\textbf{Next Item Category Prediction}} \\
\midrule
\textbf{Method} 
& \textbf{Accuracy} $\uparrow$ 
& \textbf{F1 weighted} $\uparrow$ \\
\midrule
DuoRec      & 0.4703 $\pm$ 0.0008 & 0.4521 $\pm$ 0.0010 \\
SASRec      & 0.3728 $\pm$ 0.0007 & 0.3402 $\pm$ 0.0011 \\
HSTU        & 0.3905 $\pm$ 0.0009 & 0.3575 $\pm$ 0.0020 \\
PANTHER     & 0.4038 $\pm$ 0.0017 & 0.3699 $\pm$ 0.0022 \\
Transformer & 0.4355 $\pm$ 0.0010 & 0.4127 $\pm$ 0.0012 \\
FAME        & \textbf{0.4867 $\pm$ 0.0035} & \textbf{0.4649 $\pm$ 0.0029} \\
\textbf{MoTIF (ours)}
            & \underline{0.4806 $\pm$ 0.0003}
            & \underline{0.4564 $\pm$ 0.0007} \\
\bottomrule
\end{tabular}

\vspace{0.35em}
\begin{minipage}{0.95\linewidth}
\footnotesize
\end{minipage}
\end{table}
For next item category prediction, FAME achieved the best accuracy and weighted F1, with MoTIF a close second. Temporal pattern constraints thus produced a modest asymmetry: a small gain on next amount prediction against a small cost on next item category. This is consistent with amount aligning more closely with the temporal structure MoTIF assumes, while category membership follows patterns that amount alone does not fully capture, unsurprising given that the routing supervision teacher was trained on amount alone.

\subsection{Downstream Credit Card Fraud Prediction Results}\label{appendix:cct_fraud}

\begin{table}[th]
\caption{\textbf{CCT fraud prediction, full DCN head.} Transaction-level fraud prediction from sequence embeddings and handcrafted features. $\tau^*$ is validation-tuned for precision, recall, and F1. Bold: best; underline: second-best.}
\label{tab:cct_phase2_perf}
\begin{center}
\resizebox{\textwidth}{!}{
\begin{tabular}{lccccc}
\toprule
\bf Method  & \bf AUROC & \bf AUPRC & \bf Precision@$\boldsymbol\tau^*$ & \bf Recall@$\boldsymbol\tau^*$ & \bf F1@$\boldsymbol\tau^*$\\
\hline\noalign{\smallskip}
DuoRec & 0.9554 ± .0017 & 0.8853 ± .0064 & 0.7711 ± .0150 & 0.8792 ± .0224 & 0.8212 ± .0042\\
SASRec & 0.9547 ± .0020 & 0.8826 ± .0078 & 0.7662 ± .0020 & 0.8854 ± .0077 & 0.8215 ± .0044\\
HSTU & 0.9538 ± .0027 & 0.8776 ± .0128 & 0.7727 ± .0115 & 0.8778 ± .0164 & 0.8217 ± .0022\\
PANTHER & 0.9541 ± .0014 & 0.8847 ± .0052 & 0.7651 ± .0145 & 0.8839 ± .0160 & 0.8199 ± .0031\\
Transformer & 0.9685 ± .0015 & 0.9176 ± .0042 & 0.8106 ± .0225 & 0.8953 ± .0081 & 0.8506 ± .0092\\
FAME & \textbf{0.9706 ± .0027} & \textbf{0.9271 ± .0080} & \textbf{0.8166 ± .0013} & {0.8918 ± .0067} & \textbf{0.8525 ± .0036}\\
MoTIF (ours) & \underline{0.9705 ± .0008} & \underline{0.9245 ± .0048} & \underline{0.8116 ± .0194} & \textbf{0.8962 ± .0243} & \underline{0.8513 ± .0011}\\
\bottomrule
\end{tabular}
}
\end{center}
\end{table}

\begin{table}[th]
\caption{\textbf{Embedding-only GBT fraud prediction probe.} Predicting transaction-level fraud from sequence embedding only to assess representation quality. $\tau^*$ is a threshold optimally tuned on the validation set (nested-CV) for precision, recall, and F1.Bold indicates top metric, underline indicates ours is second best.}
\label{tab:cct_phase2_perf_embedonly}
\begin{center}
\resizebox{\textwidth}{!}{
\begin{tabular}{lccccc}
\toprule
\bf Method  & \bf AUROC & \bf AUPRC & \bf Precision@$\boldsymbol\tau^*$ & \bf Recall@$\boldsymbol\tau^*$ & \bf F1@$\boldsymbol\tau^*$\\
\hline\noalign{\smallskip}
DuoRec & 0.7866 ± .0008 & 0.5702 ± .0025 & 0.4773 ± .0031 & 0.7839 ± .0116 & 0.5933 ± .0010\\
SASRec & 0.7866 ± .0020 & 0.5716 ± .0043 & 0.4844 ± .0061 & 0.7833 ± .0200 & 0.5984 ± .0016\\
HSTU & 0.7845 ± .0008 & 0.5667 ± .0012 & 0.4808 ± .0090 & 0.7874 ± .0304 & 0.5967 ± .0052\\
PANTHER & 0.7855 ± .0032 & 0.5704 ± .0046 & 0.4789 ± .0101 & \textbf{0.7942 ± .0141} & 0.5973 ± .0050\\
Transformer & 0.8309 ± .0036 & 0.6934 ± .0064 & 0.5530 ± .0136 & 0.7491 ± .0056 & 0.6362 ± .0108\\
FAME & 0.8334 ± .0050 & \textbf{0.7047 ± .0053} & \textbf{0.5570 ± .0202} & 0.7275 ± .0254 & 0.6301 ± .0048\\
MoTIF (ours) & \textbf{0.8355 ± .0037} & \underline{0.6961 ± .0041} & 0.5527 ± .0089 & \underline{0.7617 ± .0126} & \textbf{0.6405 ± .0059}\\
\bottomrule
\end{tabular}
}
\end{center}
\end{table}
To assess the generalization capability of the learned sequence representations, we evaluated their performance on a downstream fraud prediction task. For all methods, we trained a DCN head to predict fraud from the learned sequence embedding together with handcrafted features at a set of target transactions (Table~\ref{tab:cct_phase2_perf}). MoTIF and FAME performed notably better than the remaining baselines, with FAME slightly ahead on average across all metrics except recall, though overall performance is comparable. To assess representation quality more directly, we also trained an embedding-only gradient boosted tree (GBT) probe, excluding handcrafted features (Table~\ref{tab:cct_phase2_perf_embedonly}); MoTIF outperformed all baselines on F1 and AUROC while ranking second on AUPRC and recall, confirming that the learned embeddings retain strong predictive value on their own.

\end{document}